\documentclass[runningheads]{llncs}
\usepackage{graphicx}

\usepackage{tikz}
\usepackage{comment}
\usepackage{multirow}
\usepackage{makecell}
\usepackage{cite}
\usepackage{amsmath,amssymb} 
\usepackage{color}
\usepackage{hyperref}
\hypersetup{
colorlinks=true,
}

\begin{document}
\pagestyle{headings}
\mainmatter
\def\ECCVSubNumber{806}  

\title{BEVDet: High-Performance Multi-Camera \\3D Object Detection in Bird-Eye-View} 

\titlerunning{BEVDet}
%
\author{Junjie Huang \thanks{Corresponding author.}  \and Guan Huang \and Zheng Zhu\and Yun Ye \and Dalong Du}
\authorrunning{J. Huang et al.}
%
\institute{PhiGent Robotics \\
{\tt\small \email{\{junjie.huang,zhengzhu\}@ieee.org,\\ \{guan.huang, yun.ye, dalong.du\}@phigent.ai}}}

\maketitle

\begin{abstract}
Autonomous driving perceives its surroundings for decision making, which is one of the most complex scenarios in visual perception. The success of paradigm innovation in solving the 2D object detection task inspires us to seek an elegant, feasible, and scalable paradigm for fundamentally pushing the performance boundary in this area. To this end, we contribute the BEVDet paradigm in this paper. BEVDet performs 3D object detection in Bird-Eye-View (BEV), where most target values are defined and route planning can be handily performed. We merely reuse existing modules to build its framework but substantially develop its performance by constructing an exclusive data augmentation strategy and upgrading the Non-Maximum Suppression strategy. In the experiment, BEVDet offers an excellent trade-off between accuracy and time-efficiency. As a fast version, BEVDet-Tiny scores 31.2\% mAP and 39.2\% NDS on the nuScenes \texttt{val} set. It is comparable with FCOS3D, but requires just 11\% computational budget of 215.3 GFLOPs and runs 9.2 times faster at 15.6 FPS. Another high-precision version dubbed BEVDet-Base scores 39.3\% mAP and 47.2\% NDS, significantly exceeding all published results. With a comparable inference speed, it surpasses FCOS3D by a large margin of +9.8\% mAP and +10.0\% NDS. The source code is publicly available for further research\footnote {https://github.com/HuangJunJie2017/BEVDet}.

\end{abstract}

\section{Introduction}

2D visual perception has witnessed rapid development in the past few years and emerged some outstanding paradigms like Mask R-CNN \cite{Mask-RCNN}, which is high-performance, scalable\cite{CascadeRCNN, HTC}, and multi-task compatible. However, with respect to the scene of vision-based autonomous driving where both accuracy and time-efficiency are desired, major tasks like 3D object detection and map restoration (\textit{i.e.}, Bird-Eye-View (BEV) semantic segmentation) are still conducted by different paradigms in the up-to-date benchmarks. For example, in the nuScenes \cite{NS} benchmark, image-view-based methods like FCOS3D \cite{FCOS3D} and PGD \cite{PGD} have leading performances in the multi-camera 3D object detection track, while the BEV semantic segmentation track is dominated by the BEV-based methods like PON \cite{PON}, Lift-Splat-Shoot \cite{LSS}, and VPN \cite{VPN}. Which view space is more reasonable for perception in autonomous driving, and can we handle these tasks in a unified framework? Aiming at these questions, we propose BEVDet in this paper. With BEVDet, we explore the advantages of detecting 3D objects in BEV, expecting a superior performance compared to the latest image-view-based methods and a consistent paradigm with BEV semantic segmentation. In this way, we can further verify the feasibility of multi-task learning, which is meaningful for time-efficient inference.

\begin{figure*}[t]
		\centering
		\includegraphics[width=1.0\linewidth]{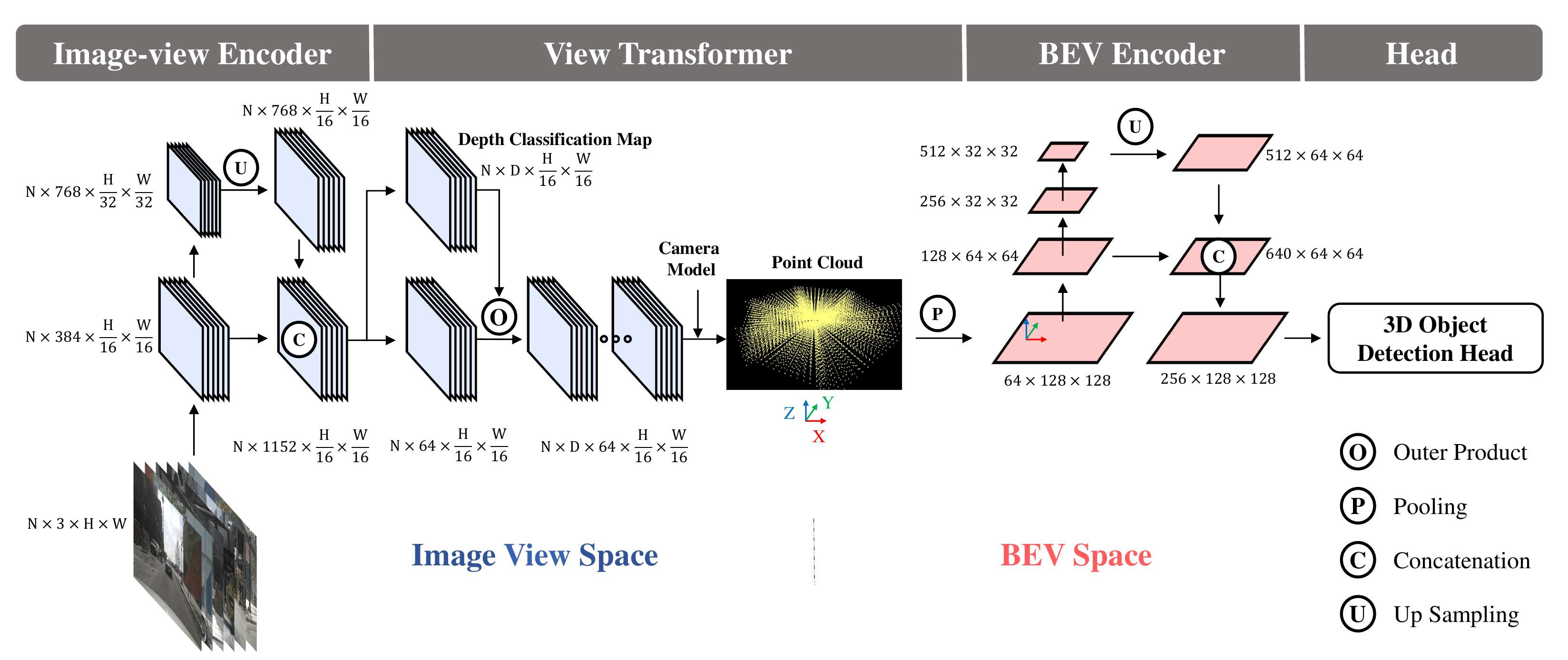}
		\caption{The framework of the proposed BEVDet paradigm. BEVDet with a modular design consists of four modules: Image-view encoder, including a backbone and a neck, is applied at first for image feature extraction. View transformer transforms the feature from the image view to BEV. BEV encoder further encodes the BEV features. Finally, a task-specific head is built upon the BVE features and predicts the target values of the 3D objects. We take BEVDet-Tiny as an example for illustrating the channels of different modules.}
		\label{fig:pipeline}
\end{figure*}

The proposed BEVDet, as illustrated in Fig.~\ref{fig:pipeline}, shares a similar framework with the up-to-date BEV semantic segmentation algorithms \cite{PON,LSS,PYVA}. It is modularly designed with an image-view encoder for encoding features in image view, a view transformer for transforming the feature from image view into BEV, a BEV encoder for further encoding features in the BEV perspective, and a task-specific head for performing 3D object detection in the BEV space. Benefiting from this modular design, we can reuse a mass of existing works which have been proved effective in other areas and still expect a long way to upgrade this paradigm specific for the task of 3D object detection.


Though simple in constructing the framework of BEVDet, it is nontrivial to build its robust performance. When verifying the feasibility of BEVDet, for reasonable performance, the data processing strategy and the parameter number of BEVDet are set close to the image-view-based 3D object detector like FCOS3D \cite{FCOS3D} and PGD \cite{PGD}. Unexpectedly, a serious over-fitting problem is observed in the training process. Some cues reveal that the devil is in the excessive fitting capacity of BEVDet in the BEV space. First of all, the over-fitting encourages us to apply a complicated data augmentation strategy in the image view space as Lift-Splat-Shoot \cite{LSS} for a regularization effect. However, this modification has a positive effect only if the BEV encoder is absent. Otherwise, it even degrades the performance. On the other hand, the batch size of the image-view encoder is N (\textit{i.e.}, the numbers of cameras like 6 in nuScenes \cite{NS}) times that of the subsequence modules. Insufficient training data is also partly responsible for the over-fitting in BEV space learning. Besides, we observe that the view transformer connects the image view space with the BEV space in a pixel-wise manner, which decouples them from the perspective of data augmentation. This makes the data augmentation in image view have no regularization effect on the subsequence modules (\textit{i.e.}, the BEV encoder and the 3D object detection head). Thus, as a supplement, additional data augmentation operations are conducted in the BEV space like flipping, scaling, and rotating for the model's robustness on these aspects. This works well in preventing BEVDet from over-fitting.



In addition, we upgrade the classical Non-Maximum Suppression (NMS) strategy for improving its adaptability in the 3D object detection scenario. The inference process is further sped up by removing the sequentially executed operators. With these modifications, BEVDet offers an outstanding trade-off between accuracy and inference latency among existing paradigms. On nuScenes \cite{NS} \texttt{val} set, the high-speed version, BEVDet-Tiny, achieves superior accuracy (\emph{i.e.}, 31.2\% mAP and 39.2\% NDS) with an image size of 704$\times$256, which is merely $1/8$ of the competitors' (\emph{i.e.}, 29.5\% mAP and 37.2\% NDS with 1600$\times$900 image size in FCOS3D \cite{FCOS3D}). Scaling down the image size reduces the computational budget by 89\% and offers a dramatic acceleration of 9.2 times (\emph{i.e.}, BEVDet with 215.3 GFLOPs and 15.6 FPS \emph{v.s.} FCOS3D with 2,008.2 GFLOPs and 1.7 FPS). By constructing another high-precision configuration dubbed BEVDet-Base, we report a new record of 39.3\% mAP and 47.2\% NDS. Moreover, compared to the existing paradigms, explicitly encoding features in BEV space makes BEVDet talented at perceiving the targets' translation, scale, orientation, and velocity. More characteristic of BEVDet can be found in ablation study.

\section{Related Works}

\subsection{Vision-based 2D Perception}
\subsubsection{Image Classification} The renaissance of deep learning for vision-based 2D perception can be dated back to AlexNet \cite{AlexNet} for image classification. From then on, the research community keeps pushing the performance boundary of image encoder by giving raise to residual \cite{ResNet}, high-resolution \cite{HRNet}, attention-based \cite{ViT}, and many other types of structures \cite{Efficientnet, Mobilenets, Res2net, DenseNet, RegNet}. And at the same time, the powerful image encoding capacity also boosts the performance of other complicated tasks like object detection \cite{FasterRCNN,RetinaNet}, semantic segmentation \cite{kirillov2020pointrend, xiao2018unified}, human pose estimation \cite{HRNet,UDP}, and so on. As a simple task, the solution pattern of image classification is dominated by Softmax \cite{AlexNet} and its derivatives. Determined by the network structure, the capacity of the image encoders plays a vital role in this problem and is the main concern in the research community.

\subsubsection{Object Detection} Common object detection, demanding both category labels and the locating bounding boxes of all pre-defined objects, is a more complicated task where paradigms also play a vital role. Two-stage method Faster R-CNN \cite{FasterRCNN}, one-stage method RetinaNet \cite{RetinaNet}, and their derivatives \cite{Mask-RCNN, CascadeRCNN, HTC, FCOS, ATSS} are the dominant methods in this area \cite{COCO, Objects365}. Inspired by Mask R-CNN \cite{Mask-RCNN}, multi-task learning has been an appealing paradigm in both the research and industry community, owing to its great potential for saving computational resources by sharing backbone and promoting tasks by training jointly. The great impact of paradigm innovation in this area inspires us to exploit superior paradigms for better perception performance in the scene of autonomous driving, where the tasks are even more complicated and multi-task learning is rather appealing.

\subsection{Semantic Segmentation in BEV}
One of the main perception tasks in autonomous driving is to vectorially restore the map of its surrounding environment. This can be achieved by semantic segmentation in BEV for the targets like drivable areas, car parking, lane dividers, stopping lines, and so on. The vision-based methods with leading performance in benchmark \cite{NS} are always with a similar framework\cite{PON,LSS,VPN,PYVA}. In this framework, there are four main components: an image-view encoder for encoding features in image view, a view transformer for transforming the features from image view to BEV, a BEV encoder for further encoding the feature in BEV, and a head for pixel-wise classification. The success of this pipeline in BEV semantic segmentation encourages us to extend it to the 3D object detection task, expecting that the features in BEV can work well in capturing some targets of 3D objects like scale, orientation, and velocity. Besides, we are also seeking a scalable paradigm in which multi-tasks learning can be achieved with both high accuracy and high efficiency.

\subsection{Vision-based 3D Object Detection}
3D object detection is another pivotal perception task in autonomous driving. In the last few years, KITTI \cite{KITTI} benchmark has fueled the rapid development of monocular 3D object detection \cite{lu2021geometry,liu2021autoshape,zhang2021objects,zou2021devil,zhou2021monocular,reading2021categorical,wang2021progressive,wang2021depth,kumar2021groomed}. However, the limited data and the single view make it incapable of developing more complicated tasks. Recently, some large-scale benchmark \cite{NS, Waymo} have been released with more data and multiple views, offering new perspectives toward the paradigm development in multi-camera 3D object detection. Based on these benchmarks, some multi-camera 3D object detection paradigms have been developed with competitive performance. For example, inspired by the success of FCOS \cite{FCOS} in 2D detection, FCOS3D \cite{FCOS3D} treats the 3D object detection problem as a 2D object detection problem and conducts perception just in image view. Benefitting from the strong spatial correlation of the targets' attribute with the image appearance, it works well in predicting this but is relatively poor in perceiving the targets' translation, velocity, and orientation. Following DETR \cite{DETR}, DETR3D \cite{DETR3D} proposes to detect 3D objects in an attention pattern, which has similar accuracy as FCOS3D. Although DETR3D requires just half the computational budget, the complex calculation pipeline slows down its inference speed to the same level as FCOS3D. PGD \cite{PGD} further develops the FCOS3D paradigm by searching and resolving with the outstanding shortcoming (\emph{i.e.} the prediction of the targets' depth). This offers a remarkable accuracy improvement on the baseline but at the cost of more computational budget and additional inference latency. The existing paradigms have limited tradeoffs between accuracy and time-efficiency. This motivates us to seek and develop new ones for substantially pushing the performance boundary in this area.

There are some pioneers \cite{reading2021categorical, PointPillar, DSGN}, who have exploited the 3D object detection task in BEV. Among them, also inspired by Lift-Splat-Shoot \cite{LSS}, \cite{reading2021categorical} is the most similar one as ours. They apply the Lift-Splat-Shoot paradigm on monocular 3D object detection and make it competitive in the KITTI \cite{KITTI} benchmark by referring to LiDARs for the supervision on depth prediction. A close idea can be found in the concurrent work of DD3D \cite{DD3D}. Differently, without the reliance on LiDARs, we upgrade this paradigm by constructing an exclusive data augmentation strategy based on the decoupling effect of the view transformer. This is a more feasible way and plays the essential role in enabling the BEVDet paradigm to perform competitively among existing methods.

\section{Methodology}

\subsection{Network Structure}
As illustrated in Fig.~\ref{fig:pipeline}, BEVDet with a modular design consists of four kinds of modules: an image-view encoder, a view transformer, a BEV encoder, and a task-specific head. We study the feasibility of BEVDet by constructing several derivatives with different structures as listed in Tab.~\ref{tab:bevdet-architecture}.

\subsubsection{Image-view Encoder} The image-view encoder encodes the input images into high-level features. To exploit the power of multi-resolution features, the image-view encoder includes a backbone for high-level feature extraction and a neck for multi-resolution feature fusion. By default, we use the classical ResNet \cite{ResNet} and the up-to-date attention-based SwinTransformer \cite{SwinTransformer} as backbone for prototype study. The substitutions include DenseNet \cite{DenseNet}, HRNet \cite{HRNet} and so on. With respect to the neck module, we use the classical FPN \cite{FPN} and the neck structure proposed in \cite{LSS}, which is named FPN-LSS in the following. FPN-LSS simply upsamples the feature with 1/32 input resolution to 1/16 input resolution and concatenates it with the one generated by the backbone. More complicated neck modules have not been exploited like PAFPN \cite{PAFPN}, NAS-FPN \cite{NAS-FPN} and so on.

\begin{table*}[t]
  \centering
  \caption{The components of BEVDet. `-number' denotes the number of channels in this module. Lift-Splat-Shoot-64-0.4$\times$0.4 denotes the view transformer proposed in \cite{LSS}. The output feature has a channel number of 64 and a resolution of 0.4 meters.}

  \resizebox{\linewidth}{!}{
  \centering
  \begin{tabular}{c|c|c|c|c}
  \hline

  \hline
  \textbf{Module}                                                & \textbf{BEVDet-Base}           & \textbf{BEVDet-Tiny}           & \textbf{BEVDet-R50}            &\textbf{BEVDet-R101}\\
  \hline
  \makecell[c]{Input \\Resolution}                      &1600$\times$640        &704$\times$256         &704$\times$256         &704$\times$256 \\
  \hline
  \multirow{2}{*}{\makecell[c]{Image-view \\ Encoder}}  & SwinTransformer-Base  &SwinTransformer-Tiny   & ResNet-50             &ResNet-101\\
  \cline{2-5}
                                                        & FPN-LSS-512           & FPN-LSS-512           & FPN-512               & FPN-256 \\
  \hline
  \makecell[c]{View\\ Transformer}                      &\makecell[c]{Lift-Splat-Shoot-64\\-0.4$\times$0.4}    & \makecell[c]{Lift-Splat-Shoot-64 \\-0.8$\times$0.8}   & \makecell[c]{Lift-Splat-Shoot-80\\-0.8$\times$0.8}   & \makecell[c]{Lift-Splat-Shoot-64\\-0.8$\times$0.8} \\
  \hline
  \multirow{4}{*}{\makecell[c]{BEV\\Encoder}}           & 2$\times$Basic-128    &2$\times$Basic-128     &2$\times$ Basic-160    &1$\times$ Basic-128\\
                                                        & 2$\times$Basic-256    &2$\times$Basic-256     &2$\times$ Basic-320    &1$\times$ Basic-256\\
                                                        & 2$\times$Basic-512    &2$\times$Basic-512     &2$\times$ Basic-640    &1$\times$ Basic-512\\
  \cline{2-5}
                                                        & FPN-LSS-512           &FPN-LSS-256             &FPN-LSS-256           &FPN-LSS-128\\
  \hline
  Head                                                  &          \multicolumn{4}{c}{CenterPoint Head \cite{CenterPoint3D} }\\
  \hline

  \hline
  \end{tabular}
  }
  \label{tab:bevdet-architecture}
\end{table*}

\subsubsection{View Transformer} The view transformer transforms the feature from image view to BEV. We apply the view transformer proposed in \cite{LSS} to construct the BEVDet prototype. The adopted view transformer takes the image-view feature as input and densely predicts the depth through a classification manner. Then, the classification scores and the derived image-view feature are used in rendering the predefined point cloud. Finally, the BEV feature can be generated by applying a pooling operation along the vertical direction (\emph{i.e.}, Z coordinate axis as illustrated in Fig.~\ref{fig:pipeline}). In practice, we extend the default range of depth prediction to $[1,60]$ meters with an interval of $1.25\times r$, where $r$ denotes the resolution of the output features.

\subsubsection{BEV Encoder} The BEV encoder further encodes the feature in the BEV space. Though the structure is similar to that of the image-view encoder with a backbone and a neck, it perceives some pivotal cues with high precision like scale, orientation, and velocity, as they are defined in the BEV space. We follow \cite{LSS} to utilize ResNet \cite{ResNet} with classical residual block to construct the backbone and combine the features with different resolutions by applying FPN-LSS.

\subsubsection{Head} The task-specific head is constructed upon the BEV feature. In common sense \cite{NS}, 3D object detection in automatic pilot aims at the position, scale, orientation, and speed of movable objects like pedestrians, vehicles, barriers, and so on. Without any modification, we directly adopt the 3D object detection head in the first stage of CenterPoint \cite{CenterPoint3D} for prototype verification and fair comparison with the LiDAR-based pipelines like PointPillar \cite{PointPillar} and VoxelNet \cite{VoxelNet}. The second refinement stage of CenterPoint has not been applied.

\subsection{The Customized Data Augmentation Strategy}
\subsubsection{The Isolated View Spaces} The view transformer \cite{LSS} transforms the feature from image view to BEV in a pixel-wise manner. Specifically, given a pixel in the image plane $\textbf{p}_{image}=[x_i,y_i,1]^T$ with a specific depth $d$, the corresponding coordinate in the 3D space is:
\begin{equation}\label{eq:img2camera}
  \textbf{p}_{camera}=\textbf{I}^{-1}(\textbf{p}_{image}*d)
\end{equation}
where $\textbf{I}$ is the $3\times3$ camera intrinsic matrix. Common data augmentation strategies with operations like flipping, cropping, and rotating can be formulated as a $3\times3$ transformation matrix $\textbf{A}$ \cite{UDP}. When a data augmentation strategy is applied on the input image (\emph{i.e.}, $\textbf{p}_{image}'=\textbf{A}\textbf{p}_{image}$), inverse transformation $\textbf{A}^{-1}$ should be applied in the view transformation \cite{LSS} to maintain the spatial consistency between the features and the targets in the BEV space:
\begin{equation}
\begin{split}
\textbf{p}_{camera}'&=\textbf{I}^{-1}(\textbf{A}^{-1}\textbf{p}_{image}'*d)=\textbf{p}_{camera}
\end{split}
\label{eq:viewtransform}
\end{equation}
According to Eq.~\ref{eq:viewtransform}, the augmentation strategy applied in image view space will not change the spatial distribution of the features in BEV space. This makes performing complicated data augmentation strategies in the image view space feasible for BEVDet.

\subsubsection{BEV Space Learning with Data Augmentation.} With respect to the learning in the BEV space, the number of data is less than that in the image view space as each sample contains multiple camera images (\emph{e.g.} each sample in the nuScenses benchmark contains 6 images \cite{NS}). The learning in the BEV space is thus prone to fall into over-fitting. As the view transformer isolates the two view spaces in the augmentation perspective, we construct another augmentation strategy specific for the regularization effect on the learning in BEV space. Following the up-to-date LiDAR-based methods \cite{CenterPoint3D, 3DSSD, SSN, Pointnet2}, common data augmentation operations in 2D space are adopted including flipping, scaling, and rotating. In practice, the operations are conducted both on the output feature of the view transformer and the 3D object detection targets to keep their spatial consistency. It is worth noting that this data augmentation strategy is built upon the precondition that the view transformer can decouple the image-view encoder from the subsequent module. This is a specific characteristic of BEVDet and may not be effective in the other methods \cite{FCOS3D, DETR3D, PGD}.

\subsection{Scale-NMS}
\label{sec:Scale-NMS}
\begin{figure}[t]
		\centering
		\includegraphics[width=1.0\linewidth]{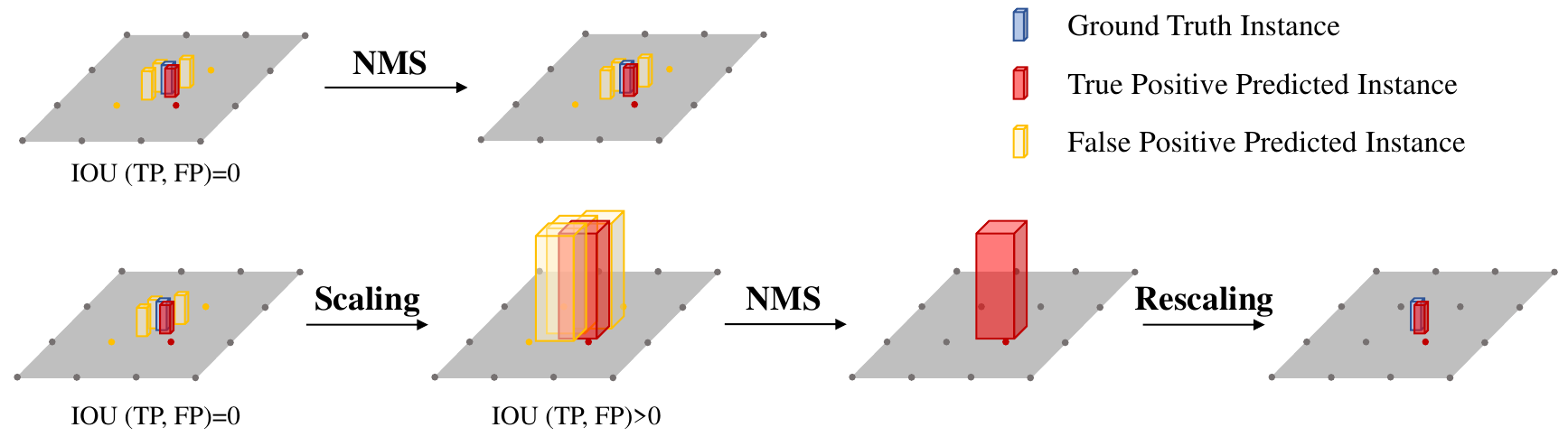}
		\caption{Illustration of the comparison between classical NMS and the proposed Scale-NMS.}
		\label{fig:nms}
\end{figure}

The spatial distribution of different categories in the BEV space is quite different from that in the image-view space. In the image-view space, all categories share a similar spatial distribution due to the perspective imaging mechanism of the cameras. Therefore, a classical Non-Maximum Suppression (NMS) \cite{NMS} strategy with a fixed threshold can work well in adjusting the predicted results of all categories to agree with the priors (\textit{e.g.}, the bounding box Intersection-Over-Union (IOU) indicator between two instances is always below a specific threshold of 0.5 in 2D object detection \cite{R-CNN, FasterRCNN, RetinaNet, FCOS, YOLO}). However, it is different in the BEV space. In the BEV space, the occupied areas of various classes are intrinsically different and the overlap between instances should be closed to zero. As a result, the distribution of IOU between predicted results varies by category. For example, as illustrated in Fig.~\ref{fig:nms}, objects like pedestrians and traffic cones occupy a small area in the ground plane, which is always smaller than the output resolution of the algorithm (\textit{e.g.}, 0.8 meters in CenterPoint \cite{CenterPoint3D}). Common object detection paradigms \cite{RetinaNet, CenterPoint3D, FCOS, FasterRCNN} redundantly generate predictions. The small occupied area of each object may make the redundant results to have no intersection with the true positive one. This deactivates the classical NMS who rely on IOU to access the spatial relationship between the true positives and the false positives.

To overcome the aforementioned problems, we proposed Scale-NMS in this paper. Scale-NMS scales the size of each object according to its category before performing the classical NMS algorithm. In this way, the distribution of IOU between the true positives and the redundant results is modulated to match up with the classical NMS. As illustrated in the second row of Fig.~\ref{fig:nms}, in predicting small objects, Scale-NMS builds the spatial relationship between results by scaling up the object size, which enables the classical NMS to drop the redundant ones according to the IOU indicator. In practice, we apply Scale-NMS to all categories except for the barrier as its size is various. The scaling factors are category-specific. They are generated by hyper-parameter searching on the validation set.

\section{Experiment}

\subsection{Experimental Settings}
\subsubsection{Dataset} We conduct comprehensive experiments on the large-scale benchmark nuScenes \cite{NS}. The nuScenes benchmark includes 1000 scenes with images from 6 cameras. It is the up-to-date popular benchmark for vision-based 3D object detection \cite{FCOS3D,DETR3D,PGD,DD3D} and BEV semantic segmentation \cite{PON,LSS,VPN,PYVA}. The scenes are officially split into 700/150/150 scenes for training/validation/testing. There are up to 1.4M annotated 3D bounding boxes for 10 classes: car, truck, bus, trailer, construction vehicle, pedestrian, motorcycle, bicycle, barrier, and traffic cone. Following CenterPoint \cite{CenterPoint3D}, we define the region of interest (ROI) within 51.2 meters in the ground plane with a resolution (\textit{i.e.}, the size of voxel in CenterPoint \cite{CenterPoint3D}) of 0.8 meters by default.

\subsubsection{Evaluation Metrics} For 3D object detection, we report the official predefined metrics: mean Average Precision (mAP), Average Translation Error (ATE), Average Scale Error (ASE), Average Orientation Error (AOE), Average Velocity Error (AVE), Average Attribute Error (AAE), and NuScenes Detection Score (NDS). The mAP is analogous to that in 2D object detection \cite{COCO} for measuring the precision and recall, but defined based on the match by 2D center distance on the ground plane instead of the Intersection over Union (IOU) \cite{NS}. NDS is the composite of the other indicators for comprehensively judging the detection capacity. The remaining metrics are designed for calculating the positive results' precision on the corresponding aspects (\emph{e.g.}, translation, scale, orientation, velocity, and attribute).

\subsubsection{Training Parameters}
Models are trained with AdamW \cite{AdamW} optimizer, in which gradient clip is exploited with learning rate 2e-4, a total batch size of 64 on 8 NVIDIA GeForce RTX 3090 GPUs. For ResNet \cite{ResNet} based image-view encoder, we apply a step learning rate policy, which drops the learning rate at epoch 17 and 20 by a factor of 0.1. With respect to SwinTransformer \cite{SwinTransformer} based image-view encoder, we apply a cyclic policy \cite{Second}, which linearly increases the learning rate from 2e-4 to 1e-3 in the first 40\% schedule and linearly decreases the learning rate from 1e-3 to 0 in the remainder epochs. By default, the total schedule is terminated within 20 epochs.

\subsubsection{Data Processing} We use $W_{in}\times H_{in}$ to denote the width and height of the input image. By default in the training process, the source images with 1600$\times$900 resolution \cite{NS} are processed by random flipping, random scaling with a range of $s\in[W_{in}/1600-0.06, W_{in}/1600+0.11]$, random rotating with a range of $r\in[-5.4^{\circ}, 5.4^{\circ}]$, and finally cropping to a size of $W_{in}\times H_{in}$. The cropping is conducted randomly in the horizon direction but is fixed in the vertical direction (\emph{i.e.}, $(y_1,y_2) =(max(0,s*900 - H_{in}), y_1+H_{in})$, where $y_1$ and $y_2$ are the upper bound and the lower bound of the target region.) In the BEV space, the input feature and 3D object detection targets are augmented by random flipping, random rotating with a range of $[-22.5^{\circ}, 22.5^{\circ}]$, and random scaling with a range of $[0.95, 1.05]$. Following CenterPoint \cite{CenterPoint3D}, all models are trained with CBGS \cite{CBGS}. In testing time, the input image is scaled by a factor of $s=W_{in}/1600+0.04$ and cropped to $W_{in}\times H_{in}$ resolution with a region defined as $(x_1,x_2,y_1,y_2)=(0.5*(s*1600-W_{in}), x_1+W_{in}, s*900-H_{in}, y_1+H_{in})$.

\subsubsection{Inference Speed} We conduct all experiments based on MMDetection3D \cite{mmdet3d2020}. All inference speeds and computational budgets are tested without data augmentation. For monocular paradigms like FCOS3D \cite{FCOS3D} and PGD \cite{PGD}, the inference speeds are divided by a factor of 6 (\emph{i.e.} the number of images in a single sample \cite{NS}), as they take each image as an independent sample. It is worth noting that, the dividing operation may not be the optimal method, as processing in the batch pattern can speed up the inference of monocular paradigms. We accelerate the proposed BEVDet paradigm by replacing the accumulative sum operation in the view transformation with another equivalent implementation. Details can be found in the ablation study section.

\begin{table*}[t]
  \centering
  \caption{Comparison of different paradigms on the nuScenes \texttt{val} set. $\dag$ initialized from a FCOS3D backbone. $\S$ with test-time augmentation. $\#$ with model ensemble.}
  	\resizebox{\linewidth}{!}{
    \begin{tabular}{l|crrc|cccccccr}
    \hline

    \hline
    Methods                     &Image Size         & \#param.  &GFLOPs     & Modality & \textbf{mAP}$\uparrow$ & mATE$\downarrow$  & mASE$\downarrow$   & mAOE$\downarrow$  & mAVE$\downarrow$  &  mAAE$\downarrow$ & \textbf{NDS}$\uparrow$   &FPS\\
    \hline
    VoxelNet \cite{CenterPoint3D}&-                 & -         &-          & LiDAR    &0.564          &-                  &-                  &-                  &-                  &-                  &0.648            &-\\
    PointPillar \cite{CenterPoint3D}&-              & -         &-          & LiDAR    &0.503          &-                  &-                  &-                  &-                  &-                  &0.602            &-\\
    \hline
    CenterNet\cite{CenterNet}   &-                  &-          &-          & Camera   & 0.306         & 0.716             & 0.264              & 0.609             & 1.426             & 0.658             & 0.328           &-\\
    FCOS3D \cite{FCOS3D}        &1600$\times$900    &52.5M      &2,008.2    & Camera   & 0.295         & 0.806             & 0.268              & 0.511             & 1.315             & \textbf{0.170}    & 0.372           &1.7\\
    DETR3D \cite{DETR3D}        &1600$\times$900    &51.3M      &1,016.8    & Camera   & 0.303         & 0.860             & 0.278              & 0.437             & 0.967             & 0.235             & 0.374           &2.0\\
    PGD \cite{PGD}              &1600$\times$900    &53.6M      &2,223.0    & Camera   & 0.335         & 0.732             & 0.263              & 0.423             & 1.285             & 0.172             & 0.409           &1.4\\
    \textbf{BEVDet-Tiny}        &704$\times$256     &53.7M      &215.3      & Camera   & 0.312         & 0.691             & 0.272              & 0.523             & 0.909             & 0.247             & 0.392           &\textbf{15.6}\\
    \textbf{BEVDet-Base}        &1600$\times$640    &126.6M     &2,962.6    & Camera   & \textbf{0.393}& \textbf{0.608}    & \textbf{0.259}     & \textbf{0.366}    & \textbf{0.822}    & 0.191             & \textbf{0.472}  &1.9\\
    \hline
    FCOS3D$\dag\S\#$ \cite{FCOS3D} &1600$\times$900  &-          &-          & Camera   & 0.343         & 0.725             & 0.263              & 0.422             & 1.292             & \textbf{0.153}    & 0.415           &-\\
    DETR3D$\dag$ \cite{DETR3D}   &1600$\times$900    &51.3M      &-          & Camera   & 0.349         & 0.716             & 0.268              & 0.379             & 0.842             & 0.200             & 0.434           &-\\
    PGD$\dag\S$ \cite{PGD}       &1600$\times$900    &53.6M      &-          & Camera   & 0.369         & 0.683             & 0.260              & 0.439             & 1.268             & 0.185             & 0.428           &-\\
    \textbf{BEVDet-Base}$\S$    &1600$\times$640    &126.6M     &-          & Camera   & \textbf{0.397}& \textbf{0.595}    & \textbf{0.257}     & \textbf{0.355}    & \textbf{0.818}    & 0.188             & \textbf{0.477}  &-\\
    \hline

    \hline
    \end{tabular}%
    }
  \label{tab:nus-val}%
\end{table*}%

\subsection{Benchmark Results}

\subsubsection{nuScenes \texttt{val} set}

We comprehensively compare the proposed BEVDet with other paradigms like FCOS3D \cite{FCOS3D}, its upgraded version PGD \cite{PGD}, and DETR3D \cite{DETR3D}. Their numbers of parameters, computational budget, inference speed, and accuracy on the nuScenes \texttt{val} set are all listed in Tab.~\ref{tab:nus-val}.

As a high-speed version dubbed BEVDet-Tiny, we set the number of parameters close to competitors and equip it with a small input resolution of 704$\times$256. With merely 1/8 input size of the competitors (\emph{i.e.}, 704$\times$256 for BEVDet-Tiny \emph{v.s.} 1600$\times$900 for FCOS3D, DETR3D, and PGD), BEVDet-Tiny requires just 215.3 GFLOPs computational budget and can be processed in 15.6 FPS. It scores 31.2\% mAP and 39.2\% NDS, which has a superior accuracy than FCOS3D (29.5\% mAP and 37.2\% NDS) and DETR3D (30.3\% mAP and 37.4\% NDS). However, it requires far less computational budget (2,008.2 GFLOPs of FCOS3D, 1,016.8 GFLOPs of DETR3D) and has a faster inference speed (1.7 FPS of FCOS3D, 2.0 FPS of DETR3D). BEVDet-Base with 1600$\times$640 input resolution requires 2962.6 GFLOPs scores 39.3\% mAP and 47.2\% NDS. With a competitive inference speed, BEVDet-Base outperforms all published results. It significantly exceeds the previous leading method PGD by a margin of +5.8\% mAP and +6.3\% NDS. It is worth noting that though the computational budget of BEVDet-Base is nearly 3 times that of DETR3D \cite{DETR3D}, BEVDet-Base can be process at a comparable speed of 1.9 FPS. The straightforward design enables BEVDet to run faster than the existing paradigms.

Considering the translation (ATE), scale (ASE), orientation (AOE), velocity (AVE), and attribute (AAE) error of the truly positive results, BEVDet works well in predicting the targets' translation, scale, orientation, and velocity, which is consistent with common sense that it is easier for an agent to capture these kinds of information in BEV. However, BEVDet performs poorly in predicting the targets' attributes when compared with the image-view-based methods like FCOS3D and PGD. We conjecture that the attribute judgment relies on the appearance cues, which is easier for agents to perceive in the image view. The combination of these two views is a promising solution to this problem, which will be studied in future work.

\begin{table*}[t]
  \centering
  \caption{Comparison with the state-of-the-art methods on the nuScenes \texttt{test} set. $\dag$ pre-train on DDAD \cite{DDAD}.}
  	\resizebox{0.92\linewidth}{!}{
    \begin{tabular}{l|c|ccccccc}
    \hline

    \hline
    Methods                     & Modality & mAP$\uparrow$ & mATE$\downarrow$   & mASE$\downarrow$  & mAOE$\downarrow$  & mAVE$\downarrow$  &  mAAE$\downarrow$ & NDS$\uparrow$ \\
    \hline
    PointPillars (Light) \cite{PointPillar}& LiDAR& 0.305    & 0.517              & 0.290             & 0.500             & 0.316             & 0.368             & 0.453          \\
    CenterFusion \cite{Centerfusion}& Camera \& Radar& 0.326& 0.631              & 0.261             & 0.516             & 0.614             & 0.115             & 0.449          \\
    CenterPoint \cite{CenterPoint3D}& Camera \& LiDAR \& Radar& 0.671 & 0.249    & 0.236             & 0.350             & 0.250             & 0.136             & 0.714          \\
    \hline
    MonoDIS \cite{MonoDIS}       & Camera   & 0.304         & 0.738              & 0.263             & 0.546             & 1.553             & 0.134             & 0.384          \\
    CenterNet \cite{CenterNet}   & Camera   & 0.338         & 0.658              & 0.255             & 0.629             & 1.629             & 0.142             & 0.400          \\
    FCOS3D   \cite{FCOS3D}      & Camera   & 0.358         & 0.690              & 0.249             & 0.452             & 1.434             & \textbf{0.124}    & 0.428          \\
    PGD      \cite{PGD}         & Camera   & 0.386         & 0.626              & 0.245             & 0.451             & 1.509             & 0.127             & 0.448          \\
    \textbf{BEVDet}             & Camera   & \textbf{0.422}& \textbf{0.529}     & \textbf{0.236}    & \textbf{0.395}    & \textbf{0.979}    & 0.152             & \textbf{0.482}\\
    \hline
    DD3D$\dag$\cite{DD3D}       & Camera   & 0.418         & 0.572              & 0.249             & 0.368             & 1.014             & 0.124             & 0.477          \\
    DETR3D$\dag$\cite{DETR3D}   & Camera   & 0.386         & 0.626              & 0.245             & 0.394             & 0.845             & 0.133             & 0.479          \\
    \hline

    \hline
    \end{tabular}%
    }
  \label{tab:nus-test}%
\end{table*}%

\subsubsection{nuScenes \texttt{test} set}
For the nuScenes \texttt{test} set, we train BEVDet-Base configuration on the \texttt{train} and \texttt{val} sets. Single model with test time augmentation is adopted. As listed in Tab.~\ref{tab:nus-test}, BEVDet ranks first on the nuScenes vision-based 3D objection leaderboard with scores of 42.2\% mAP and 48.2\% NDS, surpassing the previous leading method PGD \cite{PGD} by +3.6\% mAP and +3.4\% NDS. This has been comparable with those relied on LiDAR sensor for pre-training like DD3D \cite{DD3D} and DETR3D \cite{DETR3D}. It is also worth noting that the accuracy of vision-based BEVDet is comparable with the classical LiDAR-based method PointPillars \cite{PointPillar} (\emph{i.e.} 30.5\% mAP and 45.3\% NDS).

\begin{table*}[t]
  \centering
  \caption{Ablation study for the data augmentation strategy on the nuScenes \texttt{val} set. IDA denotes Image-view-space Data Augmentation. BDA denotes BEV-space Data Augmentation. BE denotes BEV Encoder. }
    \resizebox{1.0\linewidth}{!}{
    \begin{tabular}{c|ccc|lc|ccccccc}
    \hline

    \hline
    ID     &IDA        & BDA       &BE         & mAP-best$\uparrow$        & NDS-best$\uparrow$        & mAP$\uparrow$ & NDS$\uparrow$ & mATE$\downarrow$  & mASE$\downarrow$  & mAOE$\downarrow$  & mAVE$\downarrow$  & mAAE$\downarrow$   \\
    \hline
    A           &           &           &\checkmark &\textbf{0.230} (e4) &\textbf{0.310} (e14)& 0.174 (-5.6\%)& 0.283 (-2.7\%)&0.734              &0.343              &0.664              &1.262              &0.298          \\
    B           &\checkmark &           &\checkmark &\textbf{0.205} (e10)&\textbf{0.308} (e14)& 0.178 (-2.7\%)& 0.303 (-0.5\%)&0.738              &0.288              &0.615              &1.030              &0.217          \\
    C           &           &\checkmark &\checkmark &\textbf{0.262} (e11)&\textbf{0.357} (e14)& 0.236 (-2.6\%)& 0.348 (-0.9\%)&0.717              &0.274              &\textbf{0.514}     &0.976              &0.221          \\
    D           &\checkmark &\checkmark &\checkmark &\textbf{0.316} (e17)&\textbf{0.393} (e19)& 0.312 (\textbf{-0.4\%}) & 0.392 (\textbf{-0.1\%})
                                                                                                                                            &\textbf{0.691}     &\textbf{0.272}     &0.523              &\textbf{0.909}     &0.247          \\
    \hline
    E           &           &           &           &\textbf{0.231} (e10)&\textbf{0.307} (e10)& 0.215 (-1.6\%)& 0.306 (-0.1\%)&0.777              &0.283              &0.703              &1.111              &0.249          \\
    F           &\checkmark &           &           &\textbf{0.276} (e14)&\textbf{0.347} (e17)& 0.269 (-0.7\%)& 0.345 (-0.2\%)&0.734              &0.274              &0.673              &0.994              &0.217          \\
    G           &           &\checkmark &           &\textbf{0.253} (e12)&\textbf{0.345} (e15)& 0.224 (-2.9\%)& 0.337 (-0.8\%)&0.734              &0.281              &0.543              &0.983              &\textbf{0.211} \\
    H           &\checkmark &\checkmark &           &\textbf{0.299} (e20)&\textbf{0.373} (e20)& 0.299 (-0.0\%)& 0.373 (-0.0\%)&0.726              &0.273              &0.536              &0.950              &0.278          \\

    \hline

    \hline
    \end{tabular}%
    }
  \label{tab:abl-da}%
\end{table*}%

\subsection{Ablation Studies}
\label{sec:ablation}
\subsubsection{Data Augmentation}
With BEVDet-Tiny in Tab.~\ref{tab:bevdet-architecture}, we study how the performance of BEVDet is developed by the customized data augmentation strategy. We adopt a fixed training schedule of 20 epochs and report both the best performances during the training process and the final performances at the last epoch. By reporting and comparing the two, we analyze how the data augmentation strategy affects the performances at the saturation point and to what degree does the data augmentation alleviates the over-fitting problem. Some key factors are considered including Image-view-space Data Augmentation (IDA), BEV-space Data Augmentation (BDA), and BEV Encoder (BE). We listed the performance of different configurations in Tab.~\ref{tab:abl-da}.

As a baseline, we simply replace the head's input feature in LiDAR-based method CenterPoint \cite{CenterPoint3D} with the one generated by the view transformer proposed in \cite{LSS}. In this configuration Tab.~\ref{tab:abl-da} (A), all augmentation strategies are absent. During the training process, indicator mAP becomes saturated early at epoch 4 with 23.0\% and falls into over-fitting in the following epochs. Finally, the performance at epoch 20 is merely 17.4\% with a drop from the best by -5.6\%, which is far poor than the image-view-based method FCOS3D (29.5\%).

By applying Image-view-space Data Augmentation (IDA) in configuration Tab.~\ref{tab:abl-da} (B), the saturation of the training process is postponed to epoch 10 (20.5\%) and finally scores 17.8\%. The best performance of this configuration is even worse than the baseline (\textit{i.e.}, Tab.~\ref{tab:abl-da} (A)). In contrast, configuration Tab.~\ref{tab:abl-da} (C) with BEV-space Data Augmentation (BDA) peaks at epoch 15 with 26.2\% mAP and finally scores 23.6\% mAP at epoch 20. This surpasses the baseline by a large margin of +3.2\% mAP at the peak point. BDA plays a more important role than IDA in training BEVDet. By combining both IDA and BDA in configuration Tab.~\ref{tab:abl-da} (D), the mAP performance peaks at epoch 17 with 31.6\% and finally scores 31.2\% at epoch 20. Compared with the baseline, the combined data augmentation strategy offers a significant performance boost of +8.6\% mAP at the peak point. The performance degeneration at epoch 20 is reduced to -0.4\%. It is worth noting that, IDA has a negative impact on the performance when BDA is absent but has a positive impact on the contrary.

To study the impact of BEV Encoder (BE), we remove BE in configuration Tab.~\ref{tab:abl-da} (E, F, G, and H). Comparing configuration Tab.~\ref{tab:abl-da} (D) with (H), BE improves the BEVDet's accuracy by +1.7\% mAP, indicating that it is one of the key components in constructing the performance of BEVDet. By comparing configuration Tab.~\ref{tab:abl-da} (F) with (G), we found that IDA can offer a positive impact when BDA is absent, which is opposite when BE is present. We conjecture that the strong perception capacity of BEV Encoder can only be built upon the presence of BDA. This can be verified by comparing the best performance of configuration Tab.~\ref{tab:abl-da} (A, B, C, and D) with (E, F, G, and H) respectively.

\subsubsection{Scale-NMS}
\begin{table*}[t]
  \centering
  \caption{Ablation study for the NMS strategy on the nuScenes \texttt{val} set. }
    \resizebox{1.0\linewidth}{!}{
    \begin{tabular}{cc|ccccccccccccccccc}
    \hline

    \hline
    Methods     & mAP            & Car            &Truck          &Bus            &Trailer        &C-Vehicle      &Pedestrian     &Motorcycle     &Bicycle        &Traffic Cone   &Barrier\\
    \hline
    NMS         & 0.295          &0.512           &0.220          &0.305          &0.153          &0.069          &0.297          &\textbf{0.273} &\textbf{0.225} &0.425          &0.467 \\
    Circular-NMS& 0.298          &\textbf{0.516}  &0.210          &0.308          &0.149          &0.066          &0.295          &0.272          &0.212          &0.451          &\textbf{0.498} \\
    Scale-NMS   & \textbf{0.312} &0.512           &\textbf{0.223} &\textbf{0.313} &\textbf{0.160} &\textbf{0.072} &\textbf{0.345} &\textbf{0.273} &\textbf{0.225} &\textbf{0.500} &\textbf{0.498} \\
    \hline

    \hline
    \end{tabular}%
    }

  \label{tab:abl-nms}%
\end{table*}%
We adopt BEVDet-Tiny in Tab.~\ref{tab:bevdet-architecture} for ablation study on the NMS strategy. As shown in Tab.~\ref{tab:abl-nms}, category by category, we compare Scale-NMS with the classical NMS and the Circular-NMS proposed in CenterPoint \cite{CenterPoint3D}. The proposed Scale-NMS significantly boosts the performance on the categories with a small occupied area like pedestrians (+4.8\% AP) and traffic cones(+7.5\% AP). The other categories with relatively large sizes also benefit from Scale-NMS like buses (+0.8\% AP), trucks (+0.3\% AP), trailers (+0.7\% AP), and construct vehicles (+0.3\% AP). The overall performance mAP is thus boosted from 29.5\% to 31.2\% with an improvement of +1.7\%.

\subsubsection{Resolution}
The resolution of the signal channel is vital for BEVDet. It not only affects the accuracy of the models but also plays a key role in the computational budget and inference latency. As BEVDet involves two view spaces, two main channel resolutions are studied here: the resolution of the input image and the resolution of the BEV encoder's input features. We perform several ablation experiment in Tab.~\ref{tab:abl-res} with some typical settings. According to the results, the resolution of input image has a large impact on the accuracy. For example, BEVDet with 1408$\times$512 input size Tab.~\ref{tab:abl-res} (E) has a +4.5\% mAP superiority on that with 704$\times$256 input size Tab.~\ref{tab:abl-res} (C). It is worth noting that with the increasing of input size, the increment of the BEVDet computational budget is limited as the computational budget of the BEV encoder and heads is unchanged. Besides, a larger input size also has a consistently positive impact on predicting the targets' translation, scale, and orientation.

\begin{table*}[t]
  \centering
  \caption{Ablation study for the resolutions of BEVDet on the nuScenes \texttt{val} set. }
    \resizebox{0.92\linewidth}{!}{
    \begin{tabular}{c|cc|cc|ccccc|cr}
    \hline

    \hline
    ID        & Input Resolution    & BEV Resolution &mAP            &NDS   & mATE$\downarrow$  & mASE$\downarrow$  & mAOE$\downarrow$  & mAVE$\downarrow$  & mAAE$\downarrow$     &GFLOPs  &FPS   \\
    \hline
    A         & $704\times 256$     &0.8 Meter       &0.312          &0.392 &0.691              &0.272              &0.523              &0.909              &0.247                  & 215.3  &15.6  \\
    B         & $1056\times 384$    &0.8 Meter       &0.333          &0.410 &0.661              &0.265              &0.509              &0.886              &0.243                  & 370.5  &8.9  \\
    \hline
    C         & $704\times 256$     &0.4 Meter       &0.315          &0.410 &0.653              &0.274              &0.492              &0.851              &0.254                  & 438.4  &10.0\\
    D         & $1056\times 384$    &0.4 Meter       &0.348          &0.417 &0.644              &0.266              &0.475              &0.916              &0.264                  & 593.6  &7.1\\
    E         & $1408\times 512$    &0.4 Meter       &0.360          &0.438 &0.638              &0.266              &0.427              &0.878              &0.213                  & 824.6  &5.0\\
    \hline

    \hline
    \end{tabular}%
    }
  \label{tab:abl-res}%
\end{table*}%

With respect to the resolution of the BEV encoder's input features, it can also be regarded as the voxel size in most classical LiDAR based methods \cite{CenterPoint3D, VoxelNet}. Improving the resolution of the BEV encoder's input features can boost the accuracy of models on mAP, mATE, and mAOE indicators, but at the cost of higher computational budget and inference latency.

\begin{table*}[t]
  \centering
  \caption{Ablation study for the image-view encoder on the nuScenes \texttt{val} set. }
    \resizebox{1.0\linewidth}{!}{
    \begin{tabular}{c|c|cc|ccccc|ccc}
    \hline

    \hline
    Configuration   & Input Resolution    &mAP            &NDS   & mATE$\downarrow$  & mASE$\downarrow$  & mAOE$\downarrow$  & mAVE$\downarrow$  & mAAE$\downarrow$      &\#param.    &GFLOPs  &FPS   \\
    \hline
    BEVDet-R50      & $704\times 256$     &0.298          &0.379 &0.725              &0.279              &0.589              &0.860              &0.245                  &53.3M         & 183.8  &16.7  \\
    BEVDet-R101     & $704\times 256$     &0.302          &0.381 &0.722              &0.269              &0.543              &0.900              &0.269                  &54.1M         & 223.6  &14.3  \\
    BEVDet-Tiny   & $704\times 256$     &0.312          &0.392 &0.691              &0.272              &0.523              &0.909              &0.247                  &53.7M         & 215.3  &15.6  \\
    \hline
    BEVDet-R50      & $1056\times 384$    &0.318          &0.389 &0.718              &0.272              &0.553              &0.897              &0.258                  &53.3M         & 311.8  &11.4  \\
    BEVDet-R101     & $1056\times 384$    &0.330          &0.396 &0.702              &0.272              &0.534              &0.932              &0.251                  &54.1M         & 452.0  &9.3  \\
    BEVDet-Tiny   & $1056\times 384$    &0.333          &0.410 &0.661              &0.265              &0.509              &0.886              &0.243                  &53.7M         & 370.5  &8.9  \\
    \hline

    \hline
    \end{tabular}%
    }

  \label{tab:abl-btie}%
\end{table*}%

\subsubsection{Backbone Type in the Image-view Encoder} We study the effect of backbone type in the image-view encoder by constructing 3 derivatives of BEVDet with different structures in Tab.~\ref{tab:bevdet-architecture}. They are all constructed under the principle of containing a similar amount of parameters. As listed in Tab.~\ref{tab:abl-btie}, two input resolutions are adopted. When changing the backbone type of the image-view encoder from ResNet-R50 \cite{ResNet} into SwinTransformer-Tiny \cite{SwinTransformer} with a low input resolution of 704$\times$256, the gains are +1.4\% mAP and +1.3\% NDS (\emph{i.e.}, BEVDet-R50 with 29.8\% mAP and 37.9\% NDS \emph{v.s.} BEVDet-Tiny with 31.2\% mAP and 39.2\% NDS). BEVDet-R50 is particularly stronger in predicting the targets' velocity, while BEVDet-Tiny has superior performance in predicting the target's translation and orientation. With respect to BEVDet-R101, the gains are merely +0.4\% mAP and +0.2\% NDS on BEVDet-R50 when a small input size of 704$\times$256 is adopted. However, the gains are +1.2\% mAP and +0.7\% NDS when a larger input size of 1056$\times$384 is applied. We conjecture that a larger receptive field plays an important role in scaling up the input size.

\begin{figure}[t]
		\centering
		\includegraphics[width=0.85\linewidth]{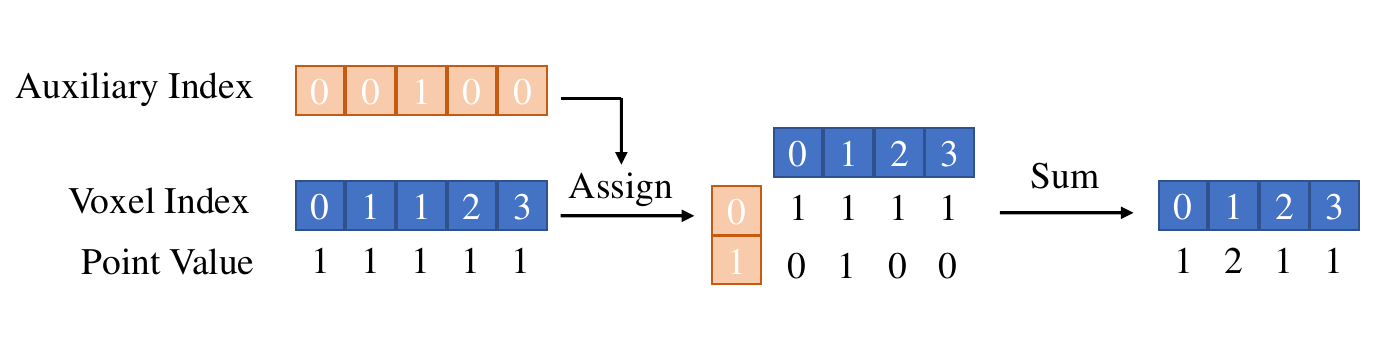}
		\caption{Combining the features with the auxiliary indexes.}
		\label{fig:speedup}
\end{figure}

\subsubsection{Acceleration} The adopted view transformation \cite{LSS} conducts accumulate sum in combining the features within the same voxel. However, the inference latency of this operation is proportional to the overall point number. To remove this operation, as illustrated in Fig.~\ref{fig:speedup}, we introduce an auxiliary index to record how many times does the same voxel index has been present previously. With this auxiliary index and the voxel index, we assign the points into a 2-D matrix and combine the features within the same voxel with a sum operation alone the auxiliary axis. Under the pre-condition that the camera intrinsic and extrinsic parameters are fixed in the inference time, the auxiliary index and the voxel index are fixed and can be calculated in the initialization phase \cite{LSS}. With this modification, we reduce the inference latency of BEVDet-Tiny by 53.3\% (\textit{i.e.}, from 137 milliseconds to merely 64 milliseconds). It is worth noting that, this modification requires extra memory which is determined by the number of voxels and the maximum value of the auxiliary index. In practice, we limit the maximum value of the auxiliary index to 300 and drop the remaining points. This operation has negligible impact on the model accuracy.

\section{Conclusion}
In this paper, we propose BEVDet, a powerful and scalable paradigm for multi-camera 3D object detection. BEVDet is constructed by referring to the success of solving semantic segmentation in BEV and is developed mainly by constructing an exclusive data augmentation strategy. In the large-scale benchmark nuSenses, BEVDet significantly pushes the performance boundary and is particularly good at predicting the targets' translation, scale, orientation, and velocity. Future works will focus on (1) improving the performance of BEVDet, particularly on targets' attribute prediction. (2) studying multi-task learning based on BEVDet.

\bibliographystyle{splncs04}
\bibliography{egbib}
\end{document}